\documentclass[sigconf,nonacm]{acmart}
\usepackage{makecell}
\usepackage{multirow}
\usepackage{enumitem}
\usepackage{color}
\usepackage[svgnames]{xcolor}
\AtBeginDocument{%
  }

\begin{document}

\title{Llama-Mob: Instruction-Tuning Llama3-8B Excels in City-Scale Mobility Prediction}

\author{Peizhi Tang}
\orcid{0009-0008-3111-172X}
\affiliation{%
  \institution{Southern University of Science and Technology}
  \country{}
}
\email{12110502@mail.sustech.edu.cn}

\author{Chuang Yang}
\orcid{0000-0002-8504-0057}
\authornote{Corresponding Author and Project Leader.}
\affiliation{%
  \institution{The Univerisity of Tokyo}
  \country{}
}
\email{chuangyang@g.ecc.u-tokyo.ac.jp}

\author{Tong Xing}
\orcid{0009-0004-5230-4652}
\affiliation{%
  \institution{The University of Tokyo}
  \country{}
}
\email{tongxing707@g.ecc.u-tokyo.ac.jp}

\author{Xiaohang Xu}
\orcid{0000-0003-1266-9943}
\affiliation{%
  \institution{The University of Tokyo}
  \country{}
}
\email{xhxu@g.ecc.u-tokyo.ac.jp}

\author{Jiayi Xu}
\orcid{0009-0003-9139-7073}
\affiliation{%
  \institution{The University of Tokyo}
  \country{}
}
\email{xujy@csis.u-tokyo.ac.jp}

\author{Renhe Jiang}
\orcid{0000-0003-2593-4638}
\authornotemark[1]
\affiliation{%
  \institution{The University of Tokyo}
  \country{}
}
\email{jiangrh@csis.u-tokyo.ac.jp}

\author{Kaoru Sezaki}
\orcid{0000-0003-1194-4632}
\affiliation{%
  \institution{The University of Tokyo}
  \country{}
}
\email{sezaki@iis.u-tokyo.ac.jp}



\renewcommand{\shortauthors}{Tang et al.}


\begin{abstract}
Human mobility prediction plays a critical role in applications such as disaster response, urban planning, and epidemic forecasting. Traditional methods often rely on designing crafted, domain-specific models, and typically focus on short-term predictions, which struggle to generalize across diverse urban environments. In this study, we introduce Llama3-8B-Mob, a large language model fine-tuned with instruction tuning, for long-term citywide mobility prediction---in a Q\&A manner. We validate our approach using large-scale human mobility data from four metropolitan areas in Japan, focusing on predicting individual trajectories over the next 15 days. The results demonstrate that Llama3-8B-Mob excels in modeling long-term human mobility—surpassing the state-of-the-art on multiple prediction metrics. It also displays strong zero-shot generalization capabilities—effectively generalizing to other cities even when fine-tuned only on limited samples from a single city. 
Moreover, our method is general and can be readily extended to the next POI prediction task. For brevity, we refer to our model as Llama-Mob, and the corresponding results are included in this paper.
Source codes are available at \underline{\url{https://github.com/TANGHULU6/Llama3-8B-Mob}}.
\end{abstract}




\maketitle

\section{Introduction}
Human mobility prediction plays a vital role in various critical scenarios, such as disaster management~\cite{jiang2018deepurbanmomentum}, epidemic forecasting~\cite{jiang2021country}, and personalized location recommendations~\cite{xu2023revisiting}. Numerous advanced machine learning models have been proposed for individual-level human mobility prediction. For instance, Feng et al. proposed a model that integrates Recurrent Neural Networks (RNNs) with attention mechanisms to jointly capture users' long- and short-term preferences for next-location prediction~\cite{feng2018deepmove}. Similarly, Xu et al. transformed both individual historical movements and collective migration patterns into graphs, applying graph learning to capture complex spatial-temporal trends, leading to significant improvements in prediction accuracy~\cite{xu2023revisiting}. However, these approaches rely on designing crafted, domain-specific models, which often struggle to generalize effectively across different cities and applications.

In recent years, Large Language Models (LLMs) have rapidly emerged, excelling not only in Natural Language Processing (NLP) tasks such as translation, sentiment analysis, and entity extraction but also redefining various scenarios like data preprocessing~\cite{jellyfish} and time series prediction~\cite{gruver2023large}. 
In this context, human mobility modeling is also experiencing a significant paradigm shift. Wang et al. designed an effective prompting strategy that enables LLMs to perform zero-shot next-location prediction in a Q\&A format for the first time~\cite{wang2023where}. The results showed that LLMs demonstrated competitive prediction performance and could offer better interpretability compared to existing domain-specific machine learning models. Furthermore, Li et al. fine-tuned the Llama-2-7B model by framing an individual’s historical trajectory as input and the next location as the response, achieving state-of-the-art (SOTA) on three real-world check-in datasets~\cite{li2024large}. These studies highlight the enormous potential of LLMs in human mobility modeling. However, several limitations remain: most work just focused on short-term predictions (i.e., next-location prediction), leaving long-term prediction largely unexplored, and the transferability among cities has yet to be examined, making it unclear whether mobility patterns learned in one city can be effectively transferred to another.

To this end, we introduce \textbf{Llama3-8B-Mob}, an instruction-tuned version of Llama3-8B~\cite{dubey2024llama}, for long-term and multi-city human mobility prediction. We validate our approach on a large-scale human trajectory dataset from four metropolitan areas in Japan, focusing on predicting human mobility over the next 15 days. The results significantly surpass the existing SOTA method, showcasing the impressive performance of Llama3-8B-Mob in \textit{long-term prediction} and its \textit{strong zero-shot generalization ability}—training in a single city effectively generalizes to others.
Beyond the above task, we further extend Llama3-8B-Mob to the next POI prediction task, termed \textbf{Llama-Mob}, and report the results in Sec. 4.5.

It is worth noting that in the Human Mobility Prediction Challenge\footnote{\url{https://wp.nyu.edu/humobchallenge2024/}} at ACM SIGSPATIAL 2024, Llama3-8B-Mob stood out among \textbf{35} models by fine-tuning with only 16\% of the training data, outperforming a range of "traditional" competitors. Specifically, it ranked \textbf{2nd} in trajectory semantic similarity (Geobleu~\cite{shimizu2022geo}), \textbf{3rd} in shape similarity (DTW~\cite{itakura1975minimum}), and had an average ranking of \textbf{1st}. This fully demonstrates the immense potential of LLMs in predicting human mobility: with a small amount of data for simple instruction fine-tuning, they can effectively mimic human movement behavior and achieve results that surpass traditional methods.



\section{Problem Definition} 

\textit{Definition 1}. (\textbf{Trajectory}): A trajectory for a moving object consists of a sequence of spatial-temporal records that depict the object's location at a specific timestamp. Formally, it is defined as:
\begin{equation}
T = (\mathit{uid}, \{(t_i, x_i, y_i) | i = 1, \dots, n\}),
\end{equation}
where \(\mathit{uid}\) is the unique identifier of the user , \((x_i, y_i)\) represents the spatial coordinates at time \(t_i\), and $n$ depicts the trajectory length.


\noindent \textit{Definition 2}. (\textbf{Individual Trajectory Prediction}): Given the past \(M\) records \(T_{t-M+1}^t\) of a trajectory \(T\), and the time information \(D_{t+1}^{t+N}\) for the future \(N\) records, the objective of individual trajectory prediction is to determine a mapping function \(f(\cdot, \cdot)\) with parameters \(\theta\), which predicts the locations of the subsequent \(N\) records:
\begin{equation} \label{eq:mobility_pred}
f(T_{t-M+1}^t, D_{t+1}^{t+N}) = \{(x_j, y_j) \mid j = t+1, \dots, t+N\},
\end{equation}
where
\begin{equation}
T_{t-M+1}^t = (\mathit{uid}, \{(t_i, x_i, y_i) \mid i = t - M + 1, \dots, t\}),
\end{equation}
and
\begin{equation}
D_{t+1}^{t+N} = \{t_j \mid j = t+1, \dots, t+N\}.
\end{equation}

\noindent \textit{Definition 3}. (\textbf{Citywide Trajectories}): Citywide trajectories, denoted as \(\mathcal{T}_{\mathcal{X}}^{\mathcal{R}}\), refer to a collection of user trajectories within a specified urban area \(\mathcal{X}\) and time range \(\mathcal{R}\).

\noindent \textit{Definition 4}. (\textbf{Citywide Trajectory Prediction}): 
Given a citywide trajectory dataset for an urban area \(\mathcal{X}\) within a past time range \(\mathcal{R}_\text{past}\), the goal is to predict the future trajectories in the \(\mathcal{X}\) over a future time range \(\mathcal{R}_\text{future}\), using the individual trajectory prediction function \(f(\cdot, \cdot)\) with parameters \(\theta\):
\begin{equation}
\mathcal{T}_{\mathcal{X}}^{\mathcal{R}_\text{past}}, \mathcal{R}_\text{future} \xrightarrow[\theta]{f(\cdot,\cdot)}  \mathcal{T}_{\mathcal{X}}^{\mathcal{R}_\text{future}}.
\end{equation}

\section{Methodology}
\subsection{Problem Reformulation}
\begin{figure}[!t] 
  \centering
  \includegraphics[width=1\linewidth]{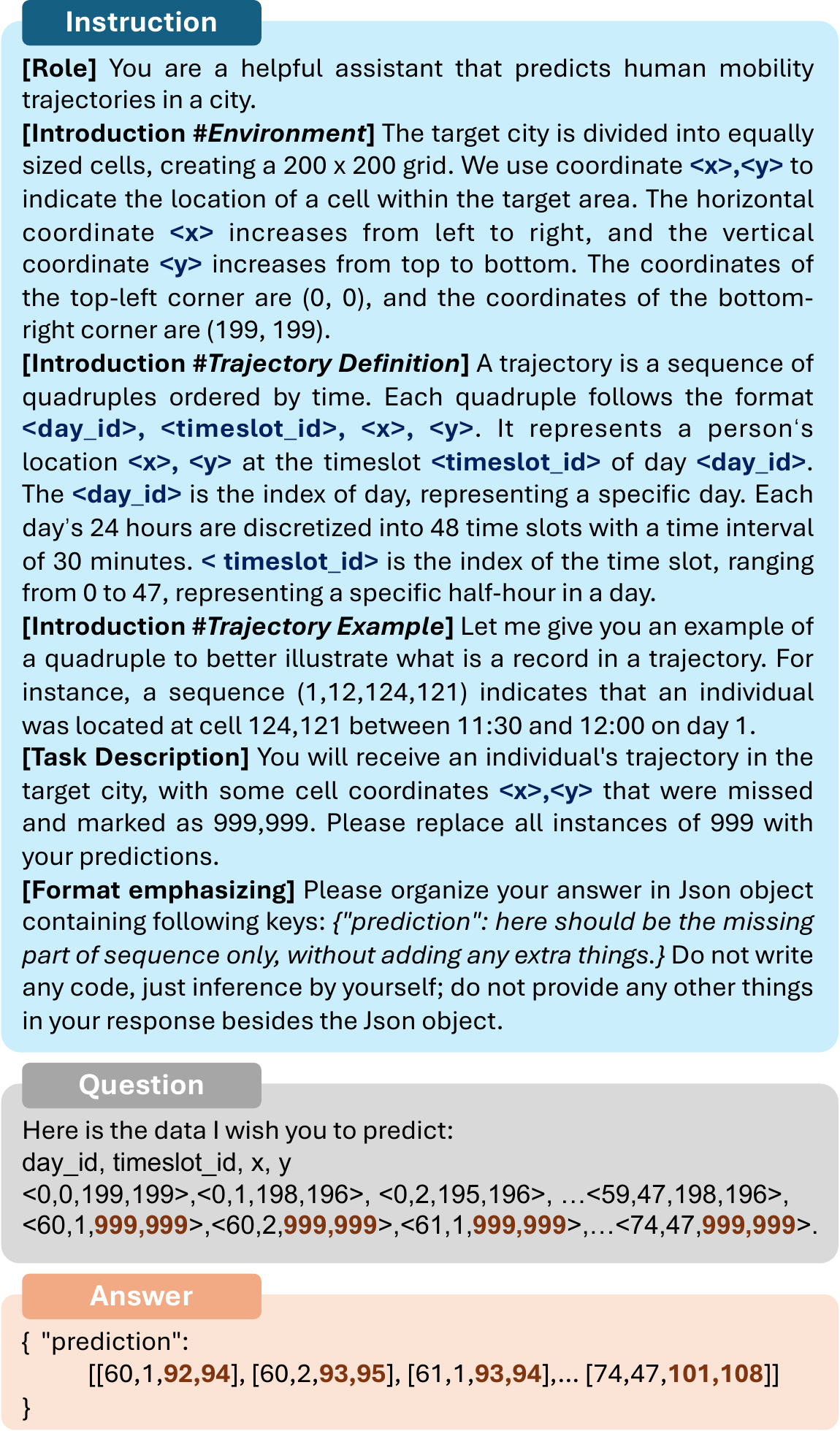}
  \caption{Framing human mobility prediction as a Q\&A task with instruction.}
  \label{figure:prompt}
\end{figure}

It is widely known that LLMs imply vast amounts of knowledge, including the common understanding of human mobility behaviors and patterns~\cite{li2024large, wang2023where}. To explore the potential of LLMs in human mobility modeling, we re-framed the trajectory prediction (Equation~\ref{eq:mobility_pred}) as a Q\&A task with instructions (Figure~\ref{figure:prompt}):
\begin{itemize}[leftmargin=*]
    \item The \textbf{instruction block} offers guidelines on the model's \textit{\#role}, \textit{\#target environment}, \textit{\#definition and example of trajectory}, \textit{\#task description}, and the \textit{\#format of the output}. 
    \item The \textbf{question block} includes the user’s \textit{\#historical trajectory} and the \textit{\#time information} for which location predictions are required. 
    \item While the \textbf{answer block} is the \textit{\#predicted future trajectory} in JSON format, which we expect as the model's response given the previous two blocks.
\end{itemize}
Compared to the \textbf{question block only} approach, this design offers LLMs a more comprehensive task context to aid in reasoning. 

\noindent $\star$ \textbf{Dataset adaptation}. 
It is important to note that the framework in Figure~\ref{figure:prompt} has been specifically adapted for the Human Mobility Challenge 2024 dataset~\cite{yabe2024yjmob100k}.

\subsection{Instruction Tuning}
\noindent \textbf{Limitations of zero-shot approach.} We conducted exploratory experiments with GPT-3.5 and 4, and found that our Q\&A framework is capable of producing decent results in a zero-shot setting. However, due to the high cost of API calls and the potential risk of data leakage in production environments, open-source models may be a better option. Nevertheless, further exploration revealed that for open-source models (i.e., the Llama series), zero-shot is not a viable solution---the models struggled to yield outputs that adhered to the desired format or aligned with the task objectives. Even when producing seemingly acceptable results occasionally, the prediction accuracy was still poor.

\noindent \textbf{Fine-tuning.}
To overcome challenges faced in zero-shot scenarios, we fine-tuned Llama3-8B~\cite{dubey2024llama} through \textit{instruction tuning}~\cite{zhang2023instruction}: 
(1) \textit{Data preparation}: we sampled a portion of users from the training set and constructed fine-tuning corpus with Figure 1 as the template, where the \textbf{instruction block} and \textbf{question block} served as the model input, and the \textbf{answer block} as the expected output. 
(2) \textit{Parameter-efficient fine-tuning}: given a large number of parameters in LLMs, we applied Low Rank Adaptation (LoRA) adapters~\cite{hu2022lora} to optimize fine-tuning efficiency, only targeting the key modules in the model, such as the query, key, value, and output projections of transformers, as well as the gate, up, and down projections. 
(3) \textit{Loss function}: we utilized token-level cross-entropy as the loss function directly, converting the complex spatial-temporal learning problem as a Q\&A fine-tuning task in the NLP domain. 

\section{Experiments} \label{sec:exp}
\subsection{Experimental Setup}
\noindent \textbf{Dataset.}
The dataset we used comes from the Human Mobility Challenge 2024~\cite{yabe2024yjmob100k}, which includes human mobility data from four metropolitan areas in Japan (cities A, B, C, and D) over a 75-day period. Each trajectory is represented as a quadruplet <x, y, day, time slot>, where x and y are discretized spatial coordinates mapped onto a 200 $\times$ 200 grid, and day and time slots represent discretized temporal information with half-hour intervals. The dataset contains complete trajectories for 100,000 individuals from city A and for 25,000, 20,000, and 6,000 individuals in cities B, C, and D, respectively. For cities B, C, and D, the location data for the last 3,000 individuals during days 61 to 75 is masked; we excluded this subset of users directly in this study. Table~\ref{tab:movement_data} shows the basic statistics of the dataset.
\begin{table}[!h]
\centering
\caption{Basic statistics of dataset.}
\small
\begin{tabular}{c|cccc}
\hline  
\multirow{2}{*}{\textbf{\makecell[c]{Attribute}}}            & \multicolumn{4}{c}{\textbf{City}} \\ 
 & \textit{A} & \textit{B} & \textit{C} & \textit{D} \\ \hline
 \textbf{Time Coverage}    & \multicolumn{4}{c}{75 days}                                   \\ \hline
 
\textbf{Time Granularity}   & \multicolumn{4}{c}{30 minutes (48 time slots per day)}                          \\ \hline
\textbf{Spatial Granularity}            & \multicolumn{4}{c}{500 meters $\times$ 500 meters (200 $\times$ 200 grids)}    \\ \hline
\textbf{\# Trajectories}    & 100,000 & 22,000 & 17,000 & 3,000                     \\ \hline
\textbf{Avg. Traj. Len.}      & 1115.35 & 998.91 & 948.08 & 1450.83 \\ \hline

\end{tabular}
\label{tab:movement_data}
\end{table}
The goal is to infer the future trajectories of cities B, C, and D between days 61-75 using the past trajectories from days 1-60, corresponding to parameters $\mathcal{R}_{past}$ and $\mathcal{R}_{future}$ as defined in \textit{Definition 4}.
To evaluate the model performance, for each city, the data is further split into training and evaluation sets in an 8:2 ratio by user ids. 
Moreover, given the efficiency constraints of LLMs in inference tasks---where predicting a single user's trajectory takes approximately 5 minutes---we randomly selected 100 users from the validation set to form a smaller evaluation subset for performance analysis within a restricted timeframe.

\noindent \textbf{Baselines.}
We used the \textbf{champion} model~\cite{terashima2023human} of the Human Mobility Prediction Challenge 2023,
namely \textbf{LP-Bert}, as our baseline. This model treats the trajectory prediction task as a BERT-based imputation task, where each location record's full information is represented as a token, with a spatial-temporal embedding layer converting it into token embedding. Since the model does not support cross-city predictions, we modified the output layer by adding a city embedding module to enable the multi-city scenario.

\noindent \textbf{Parameter Settings.}
Our base model is an optimized version of Llama3-8B, which is 2-5x faster than the original. It uses 4-bit quantization to reduce GPU memory usage by 70\%. Readers can access it here\footnote{\url{https://huggingface.co/unsloth/Llama3-8b-bnb-4bit}}. During fine-tuning, the rank of the LoRA adapters was set to 16. The training batch size was 1, with gradient accumulation over 4 steps and a learning rate of 2e-3. The number of fine-tuning epochs was set to 3. AdamW 
optimizer was used with a cosine learning rate scheduler and a weight decay of 0.01.

\noindent \textbf{Environment.}
Experiments were conducted on a linux server equipped with 2 Intel(R) Xeon(R) Silver 4310 CPUs and 4 NVIDIA RTX A6000 GPUs, each offering 48 GB of memory. All training and inference procedures were tracked and logged using the Wandb tool\footnote{\url{https://github.com/wandb/wandb}}, ensuring reproducibility and consistent reporting.

\noindent \textbf{Evaluation Metrics.}
We used Dynamic Time Warping (DTW)~\cite{itakura1975minimum} and GEO-BLEU~\cite{shimizu2022geo} to evaluate the quality of predictions. DTW measures the shape similarity between predicted and ground truth trajectories, while GEO-BLEU, a geospatial variant of BLEU, incorporates spatial proximity into n-gram matching to assess the similarity of geospatial sequences. 


\subsection{Effectiveness Evaluation}
In this section, we assess the effectiveness of LP-Bert and Llama3-8B-Mob. LP-Bert was trained using the complete training data of cities A, B, C, and D. In contrast, training Llama3-8B-Mob with the entire dataset would be impractical and time-consuming, potentially taking several months (see Section~\ref{sec:eff_eval}). Therefore, we conducted instruct tuning separately with the training data from cities B, C, and D, resulting in three distinct versions of Llama3-8B-Mob. Table~\ref{tab:combined_metrics} shows the performance of these models on the validation set. Notably, even with simple fine-tuning based solely on single city data, our model (all three variants) significantly outperforms last year's leading model which was trained on data from all cities.
We attribute this superiority to Llama3-8B's inherent understanding of human mobility patterns as well as its strong reasoning ability.
Moreover, Llama3-8B-Mob \textit{w/ single city} performs remarkably well in predictions for other cities, showcasing its generalization ability across different urban environments without the need for city-specific data! Furthermore, it can be found that fine-tuning with City B yields the best results (mean rank = 2.5). However, the average trajectory length of City B is 998.91, which may not encompass some long trajectory scenarios. Therefore, we randomly selected 1,000 trajectories from A (average trajectory length to 1115.35) to fine-tune together with B, resulting in improved outcomes: all metrics achieved SOTA results except for DTW@City-B.

\begin{table}[!h] 
\setlength \tabcolsep{3pt}
\centering
\caption{Model performance comparison.}
\label{tab:combined_metrics}
\small
\resizebox{1\linewidth}{!}{
\begin{tabular}{c|ccc|ccc|c|c}
 \hline  
  \multirow{2}{*}{\textbf{Model}} & \multicolumn{3}{c|}{\textbf{Average DTW ($\downarrow$)}} & \multicolumn{3}{c|}{\textbf{Average GEO-BLEU ($\uparrow$)}} & 
  \multirow{2}{*}{\textbf{\makecell[c]{\# Trajs Used \\ for Train}}} & \multirow{2}{*}{\textbf{\makecell[c]{Mean \\
  Rank}}} \\
  & \textit{B} & \textit{C} & \textit{D} & \textit{B} & \textit{C} & \textit{D} & & \\
\hline 
 LP-Bert~\cite{terashima2023human} & \textbf{23.30} & 23.81 & 38.89 & 0.3093 & 0.2682 & 0.3033 & 113,600 & 4.17\\
 Llama3-8B-Mob \textit{w/ B}  & 26.32 & 22.49  & \underline{34.41} & 0.3322 & \underline{0.2895} & \underline{0.3157} & 17,600  & \underline{2.50 } \\
 Llama3-8B-Mob \textit{w/ C} & 31.58 & 23.75 & 34.49 & \underline{0.3399} & 0.2891 & 0.2833 & 13,600 & 3.67 \\
 Llama3-8B-Mob \textit{w/ D} & 28.75 & \underline{22.20} & 38.46 & 0.3251 & 0.2765 & 0.3056 & 2,400  & 3.50 \\
 Llama3-8B-Mob \textit{w/ A+B} & \underline{25.39} & \textbf{20.57} & \textbf{31.94} & \textbf{0.3541} & \textbf{0.2969} & \textbf{0.3217} & 18,600 & \textbf{1.17} \\
\hline 
\end{tabular}
}
\end{table}

Due to limitations in computational resources and time, we only investigated four data fine-tuning settings here (B, C, D, A+B). Even though, we believe that there exists an optimal data combination that covers more scenarios and corner cases while minimizing information redundancy.

\subsection{Efficiency Evaluation} 
Although Llama3-8B-Mob demonstrates strong predictive capabilities, we must acknowledge that its training and inference costs are significantly higher compared to LP-Bert. Table~\ref{tab:computation_efficiency} provides an overview of the training and inference efficiency of both models. As shown, the training time for Llama3-8B-Mob \textit{w/ A+B} reaches 6.64 days, which is 2.4x longer than that of LP-Bert. Additionally, the average inference time per trajectory (batch size = 1) is 225.61 seconds—\textbf{16,000x} slower than LP-Bert. Due to the auto-regressive nature of the Llama3-8B, the inference time increases linearly with the length of the trajectory, with the longest inference time in the validation set reaching up to 15 minutes. This presents a significant challenge when applying LLMs to long-term trajectory prediction. 
\label{sec:eff_eval}
\begin{table}[!h]
\centering
\caption{Training and inference efficiency.}
\label{tab:computation_efficiency}
\footnotesize
\resizebox{1\linewidth}{!}{
\begin{tabular}{c|c|cc|cc}
 \hline  
 \multirow{2}{*}{\textbf{Model}} & \multirow{2}{*}{\textbf{{\makecell[c]{\# Trainable \\ Parameters}}}} & \multicolumn{2}{c|}{\textbf{Training}} & \multicolumn{2}{c}{\textbf{Inference}} \\ 
  &  & \textit{GPU Usage} & \textit{$t_{total}$} & \textit{GPU Usage} & \textit{$t_{infer}$} \\
\hline 
 LP-Bert~\cite{terashima2023human} & 12.20 M & 25.97 GiB & 2.77 d & 1.49 GiB & 13.94 ms \\
 Llama3-8B-Mob \textit{w/ A+B} & 41.94 M & 43.11 GiB & 6.64 d & 14.86 GiB & 225.61 s \\
\hline 
\end{tabular}
}
\end{table}

\subsection{Case Study}

\begin{figure}[!h]
    \centering
    \includegraphics[width=\linewidth]{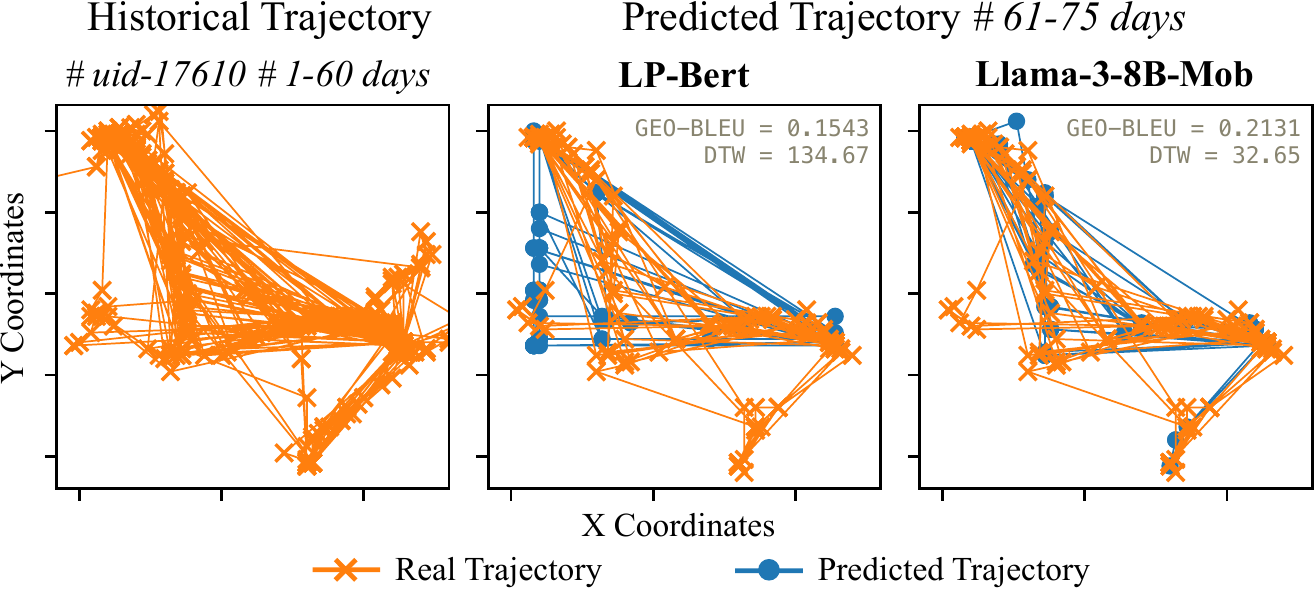}
    \caption{Visualization comparison of prediction results.}
    \label{fig:prediction_case}
\end{figure}

We conducted a case study on a randomly selected individual from city B, as shown in Figure~\ref{fig:prediction_case}. The left panel displays the user's historical trajectory for days 1-60, while the middle and right panels show the predictions from LP-Bert and Llama3-8B, respectively. As observed, LP-Bert tends to consistently predict trajectories with right-angled triangular shapes, which clearly deviate from general human mobility patterns. In contrast, Llama3-8B accurately replicates an individual’s movement behavior---the orange and blue lines nearly overlap. Notably, LP-Bert's behavior is not an isolated case; in many other examples, it also tends to predict regular geometric shapes (e.g., squares). 
Additionally, the gray text shows the DTW and GEO-BLEU scores for both models, indicating that Llama3-8B-Mob significantly outperforms LP-Bert. 

\subsection{Extension to Next POI Prediction Task}
\textit{Definition 5.} (\textbf{Next POI Prediction}): A Point of Interest (POI) refers to a specific geographic location with distinct semantic attributes. Let $\mathcal{P} = \{p_1, \ldots, p_{|\mathcal{P}|}\}$ denote the complete set of candidate POIs. Given the past $t$ check-in records $\mathbf{s}_u = (q_1, \dots, q_t)$ of a user $u$, where each record $q_k = (p_k, c_k)$ depicts the visited POI $p_k \in \mathcal{P}$ and its context $c_k$. Specifically, here the context is formulated as a tuple $c_k = (\tau_k, l_k, v_k)$, which encapsulates the check-in timestamp $\tau_k$, the geographical coordinates $l_k$, and the semantic category $v_k$ of the POI. The objective of Next-POI prediction is to determine a mapping function $f(\cdot)$ with parameters $\theta$, which predicts the subsequent POI $p_{t+1}$:
\begin{equation}
\hat{p}_{t+1} = \arg\max_{p \in \mathcal{P}} f(\mathbf{s}_u).
\end{equation}

\noindent \textbf{Datasets.}
We conduct our experiments on three public real-world datasets: Foursquare-NYC, Foursquare-TKY~\cite{yang2014modeling}, and Gowalla-CA~\cite{cho2011friendship}. These location-based social media datasets include check-ins with spatio-temporal contexts $c_k$. Following GETNext~\cite{yang2022getnext} and STHGCN~\cite{yan2023spatio}, we partition each user's check-ins to form the historical trajectory $\mathbf{s}_u$ and extract the target subsequent POI $p_{t+1}$, consistent with the sequence and prediction objective in \textit{Definition 5}.

\noindent \textbf{Evaluation Metrics.}
To evaluate the mapping function $f(\cdot)$ formulated in \textit{Definition 5}, we report Top-$K$ Accuracy (Acc@$K$) and Mean Reciprocal Rank (MRR). Specifically, Acc@$K$ measures the proportion of test instances where the ground-truth POI $p_{t+1}$ appears in the top-$K$ candidate list generated by $f(\mathbf{s}_u)$, while MRR computes the average of the reciprocal ranks of $p_{t+1}$ across all predictions.

\noindent \textbf{Baselines.}
We compare our method against traditional methods for sequential transitions like FPMC~\cite{rendle2010factorizing} and PRME~\cite{feng2015personalized}; RNN-based models including LSTM~\cite{hochreiter1997long}, ST-RNN~\cite{liu2016predicting},  DeepMove~\cite{feng2018deepmove}, STGN~\cite{zhao2019where}, LSTPM~\cite{sun2020where}, STGCN~\cite{zhao2022where}, and PLSPL~\cite{wu2022personalized}; attention-based models including STAN~\cite{luo2021stan} and MHSA~\cite{hong2023context}; graph-based models including GETNext~\cite{yang2022getnext} and STHGCN~\cite{yan2023spatio}. 
Results are listed in Table~\ref{tab:nextloc_benchmark}, where most results are inherited from GETNext~\cite{yang2022getnext} and STHGCN~\cite{yan2023spatio}, and the results of DeepMove~\cite{feng2018deepmove}, LSTPM~\cite{sun2020where}, and MHSA~\cite{hong2023context} are reproduced by us.

\noindent \textbf{Discussions.}
Our method, abbreviated as \textbf{\underline{Llama-Mob}} from Llama3-8B-Mob, attains competitive performance in NYC, while leaving room for further improvement in TKY and CA.
While our instruction-tuned LLM successfully captures trajectory semantics, the absence of explicit spatial inductive biases constrains its precision. Nevertheless, these results establish our model as a solid, simple, and general baseline for future work.
Overall, these findings indicate that our instruction-tuned framework can be effectively extended to the next POI prediction setting.
\vspace{-3mm}

\begin{table*}[!t]
\centering
\caption{Next-location prediction performance comparison (Acc@K and MRR).}
\label{tab:nextloc_benchmark}
\footnotesize
\setlength{\tabcolsep}{4pt}
\resizebox{\textwidth}{!}{
\begin{tabular}{l|cccc|cccc|cccc}
\hline
\multirow{2}{*}{\textbf{Model}} & \multicolumn{4}{c|}{\textbf{NYC}} & \multicolumn{4}{c|}{\textbf{TKY}} & \multicolumn{4}{c}{\textbf{CA}} \\
 & Acc@1 & Acc@5 & Acc@10 & MRR & Acc@1 & Acc@5 & Acc@10 & MRR & Acc@1 & Acc@5 & Acc@10 & MRR \\
\hline
FPMC~\cite{rendle2010factorizing} & 0.1003 & 0.2126 & 0.2970 & 0.1701 & 0.0814 & 0.2045 & 0.2746 & 0.1344 & 0.0383 & 0.0702 & 0.1159 & 0.0911 \\
PRME~\cite{feng2015personalized} & 0.1159 & 0.2236 & 0.3105 & 0.1712 & 0.1052 & 0.2278 & 0.2944 & 0.1786 & 0.0521 & 0.1034 & 0.1425 & 0.1002 \\
LSTM~\cite{hochreiter1997long} & 0.1305 & 0.2719 & 0.3283 & 0.1857 & 0.1335 & 0.2728 & 0.3277 & 0.1834 & 0.0665 & 0.1306 & 0.1784 & 0.1201 \\
ST-RNN~\cite{liu2016predicting} & 0.1483 & 0.2923 & 0.3622 & 0.2198 & 0.1409 & 0.3022 & 0.3577 & 0.2212 & 0.0799 & 0.1423 & 0.1940 & 0.1429 \\
DeepMove~\cite{feng2018deepmove} & 0.1520 & 0.3349 & 0.4030 & 0.2334 & 0.1733 & 0.3378 & 0.3991 & 0.2498 & 0.0928 & 0.1844 & 0.2288 & 0.1384 \\
STGN~\cite{zhao2019where} & 0.1716 & 0.3381 & 0.4122 & 0.2598 & 0.1689 & 0.3391 & 0.3848 & 0.2422 & 0.0810 & 0.1842 & 0.2579 & 0.1675 \\
LSTPM~\cite{sun2020where} & 0.2034 & 0.4897 & 0.6086 & 0.3318 & 0.2270 & 0.4815 & 0.5783 & 0.3437 & 0.1336 & 0.3110 & 0.3873 & 0.2172 \\
STGCN~\cite{zhao2022where} & 0.1799 & 0.3425 & 0.4279 & 0.2788 & 0.1716 & 0.3453 & 0.3927 & 0.2504 & 0.0961 & 0.2097 & 0.2613 & 0.1712 \\
PLSPL~\cite{wu2022personalized} & 0.1917 & 0.3678 & 0.4523 & 0.2806 & 0.1889 & 0.3523 & 0.4150 & 0.2542 & 0.1072 & 0.2278 & 0.2995 & 0.1847 \\
STAN~\cite{luo2021stan} & 0.2231 & 0.4582 & 0.5734 & 0.3253 & 0.1963 & 0.3798 & 0.4464 & 0.2852 & 0.1104 & 0.2348 & 0.3018 & 0.1869 \\
MHSA~\cite{hong2023context} & 0.1999 & 0.4968 & \textbf{0.6317} & 0.3329 & \underline{0.2417} & \underline{0.5061} & \textbf{0.6083} & \underline{0.3621} & 0.1391 & \underline{0.3218} & \underline{0.4034} & \underline{0.2281} \\
GETNext~\cite{yang2022getnext} & 0.2435 & \underline{0.5089} & 0.6143 & \underline{0.3621} & 0.2254 & 0.4417 & 0.5287 & 0.3262 & 0.1357 & 0.2852 & 0.3590 & 0.2103 \\
STHGCN~\cite{yan2023spatio} & \textbf{0.2734} & \textbf{0.5361} & \underline{0.6244} & \textbf{0.3915} & \textbf{0.2950} & \textbf{0.5207} & \underline{0.5980} & \textbf{0.3986} & \textbf{0.1730} & \textbf{0.3529} & \textbf{0.4191} & \textbf{0.2558} \\
\textbf{Llama-Mob} & \underline{0.2519} & 0.4990 & 0.5362 & 0.3507 & 0.2251 & 0.4169 & 0.4703 & 0.3059 & \underline{0.1446} & 0.2730 & 0.3198 & 0.1988 \\
\hline
\end{tabular}
}
\end{table*}

\section{Conclusion}
This study proposes an LLMs instruction tuning approach for long-term human mobility prediction across multiple cities. We validated the effectiveness and the superiority of the proposed method using large-scale human mobility data from four metropolitan areas in Japan. In the future, we will improve our work from three aspects: (1) Recognizing that the effectiveness of fine-tuning LLMs is highly dependent on data quality, \textit{designing better data selection strategies} (currently random) towards more efficient and effective fine-tuning will be next focus. (2) The slow inference time greatly affects the practicality of the Llama3-8B-Mob. We will continue to explore the latest advances in \textit{efficient inference techniques} and apply them to our model. (3) Expanding validation to \textit{more trajectory datasets}, beyond human trajectories and grid-level data.




\begin{acks}
This work was supported by JST SPRING JPMJSP2108, JSPS KAKENHI JP24K02996, and JST CREST JPMJCR21M2 including AIP challenge program, Japan.
\end{acks}

\bibliographystyle{ACM-Reference-Format}
\bibliography{sample-base}










\end{document}